\documentclass[conference]{IEEEtran}
\IEEEoverridecommandlockouts
\usepackage[maxbibnames=50,autopunct=true,babel=hyphen,hyperref=true,abbreviate=false,backend=biber]{biblatex}
\usepackage{multirow}
\usepackage{tabularx}
\usepackage{amsmath,amssymb,amsfonts}
\usepackage{booktabs}
\usepackage{algorithmic}
\usepackage{graphicx}
\usepackage{textcomp}
\usepackage[T5]{fontenc}
\usepackage[utf8]{inputenc}
\usepackage{tabularx,ragged2e}
\usepackage{pifont}
\usepackage{float}
\usepackage{caption}
\usepackage{balance}
\usepackage[colorinlistoftodos]{todonotes}
\usepackage[export]{adjustbox}
\usepackage{subcaption}
\DeclareCaptionFont{eightpt}{\fontsize{8pt}{10pt}\selectfont #1}
\usepackage[tablename=TABLE]{caption}
\usepackage{mathabx}
\usepackage{graphicx}
\usepackage{xcolor}
\usepackage{colortbl}
\usepackage{amsmath}
\usepackage{hyperref}

\bibliography{main}

\usepackage{pifont}%http://ctan.org/pkg/pifont
\newcommand{\cmark}{\ding{51}}%
\newcommand{\xmark}{\ding{55}}%

\def\BibTeX{{\rm B\kern-.05em{\sc i\kern-.025em b}\kern-.08em
    T\kern-.1667em\lower.7ex\hbox{E}\kern-.125emX}}

\begin{document}

\title{
%Automatic Detection of Maps Missing Designated Landmarks using Computer Vision
Detecting Omissions in Geographic Maps through Computer Vision
}

\author{\IEEEauthorblockN{Phuc Nguyen}
\IEEEauthorblockA{\textit{VinAI Research, Vietnam} \\
v.phucnda@vinai.io}
\and
\IEEEauthorblockN{Anh Do}
\IEEEauthorblockA{\textit{Ministry of Information \& Communications, } \\ \textit{Vietnam}, dcanh@mic.gov.vn}
\and
\IEEEauthorblockN{Minh Hoai}
\IEEEauthorblockA{\textit{VinAI Research, Vietnam} \\
v.hoainm@vinai.io} 
}
% \and
% \and
% \and

\maketitle

\begin{abstract}
This paper explores the application of computer vision technologies to the analysis of maps, an area with substantial historical, cultural, and political significance. Our focus is on developing and evaluating a method for automatically identifying maps that depict specific regions and feature landmarks with designated names, a task that involves complex challenges due to the diverse styles and methods used in map creation. We address three main subtasks: differentiating maps from non-maps, verifying the accuracy of the region depicted, and confirming the presence or absence of particular landmark names through advanced text recognition techniques. Our approach utilizes a Convolutional Neural Network and transfer learning to differentiate maps from non-maps, verify the accuracy of depicted regions, and confirm landmark names through advanced text recognition. We also introduce the VinMap dataset, containing annotated map images of Vietnam, to train and test our method. Experiments on this dataset demonstrate that our technique achieves F1-score of 85.51\% for identifying maps excluding specific territorial landmarks. This result suggests practical utility and indicates areas for future improvement. \url{https://github.com/VinAIResearch/VinMap}

%Our contributions include a novel dataset and a computer vision method that underscores the potential for further research in automated map analysis, supporting applications in various fields.

\end{abstract}


\def\mA{\mathcal{A}}
\def\mB{\mathcal{B}}
\def\mC{\mathcal{C}}
\def\mD{\mathcal{D}}
\def\mE{\mathcal{E}}
\def\mF{\mathcal{F}}
\def\mG{\mathcal{G}}
\def\mH{\mathcal{H}}
\def\mI{\mathcal{I}}
\def\mJ{\mathcal{J}}
\def\mK{\mathcal{K}}
\def\mL{\mathcal{L}}
\def\mM{\mathcal{M}}
\def\mN{\mathcal{N}}
\def\mO{\mathcal{O}}
\def\mP{\mathcal{P}}
\def\mQ{\mathcal{Q}}
\def\mR{\mathcal{R}}
\def\mS{\mathcal{S}}
\def\mT{\mathcal{T}}
\def\mU{\mathcal{U}}
\def\mV{\mathcal{V}}
\def\mW{\mathcal{W}}
\def\mX{\mathcal{X}}
\def\mY{\mathcal{Y}}
\def\mZ{\mathcal{Z}} 

\def\bbN{\mathbb{N}} 
\def\bbR{\mathbb{R}} 
\def\bbP{\mathbb{P}} 
\def\bbQ{\mathbb{Q}} 
\def\bbE{\mathbb{E}}

\def\1n{\mathbf{1}_n}
\def\0{\mathbf{0}}
\def\1{\mathbf{1}}

\def\A{{\bf A}}
\def\B{{\bf B}}
\def\C{{\bf C}}
\def\D{{\bf D}}
\def\E{{\bf E}}
\def\F{{\bf F}}
\def\G{{\bf G}}
\def\H{{\bf H}}
\def\I{{\bf I}}
\def\J{{\bf J}}
\def\K{{\bf K}}
\def\L{{\bf L}}
\def\M{{\bf M}}
\def\N{{\bf N}}
\def\O{{\bf O}}
\def\P{{\bf P}}
\def\Q{{\bf Q}}
\def\R{{\bf R}}
\def\S{{\bf S}}
\def\T{{\bf T}}
\def\U{{\bf U}}
\def\V{{\bf V}}
\def\W{{\bf W}}
\def\X{{\bf X}}
\def\Y{{\bf Y}}
\def\Z{{\bf Z}}

\def\a{{\bf a}}
\def\b{{\bf b}}
\def\c{{\bf c}}
\def\d{{\bf d}}
\def\e{{\bf e}}
\def\f{{\bf f}}
\def\g{{\bf g}}
\def\h{{\bf h}}
\def\i{{\bf i}}
\def\j{{\bf j}}
\def\k{{\bf k}}
\def\l{{\bf l}}
\def\m{{\bf m}}
\def\n{{\bf n}}
\def\o{{\bf o}}
\def\p{{\bf p}}
\def\q{{\bf q}}
\def\r{{\bf r}}
\def\s{{\bf s}}
\def\t{{\bf t}}
\def\u{{\bf u}}
\def\v{{\bf v}}
\def\w{{\bf w}}
\def\x{{\bf x}}
\def\y{{\bf y}}
\def\z{{\bf z}}

\def\balpha{\mbox{\boldmath{$\alpha$}}}
\def\bbeta{\mbox{\boldmath{$\beta$}}}
\def\bdelta{\mbox{\boldmath{$\delta$}}}
\def\bgamma{\mbox{\boldmath{$\gamma$}}}
\def\blambda{\mbox{\boldmath{$\lambda$}}}
\def\bsigma{\mbox{\boldmath{$\sigma$}}}
\def\btheta{\mbox{\boldmath{$\theta$}}}
\def\bomega{\mbox{\boldmath{$\omega$}}}
\def\bxi{\mbox{\boldmath{$\xi$}}}
\def\bnu{\mbox{\boldmath{$\nu$}}}                                  
\def\bphi{\mbox{\boldmath{$\phi$}}}
\def\bmu{\mbox{\boldmath{$\mu$}}}

\def\bDelta{\mbox{\boldmath{$\Delta$}}}
\def\bOmega{\mbox{\boldmath{$\Omega$}}}
\def\bPhi{\mbox{\boldmath{$\Phi$}}}
\def\bLambda{\mbox{\boldmath{$\Lambda$}}}
\def\bSigma{\mbox{\boldmath{$\Sigma$}}}
\def\bGamma{\mbox{\boldmath{$\Gamma$}}}
                                  
\newcommand{\myprob}[1]{\mathop{\mathbb{P}}_{#1}}

\newcommand{\myexp}[1]{\mathop{\mathbb{E}}_{#1}}

\newcommand{\mydelta}[1]{1_{#1}}

\newcommand{\myminimum}[1]{\mathop{\textrm{minimum}}_{#1}}
\newcommand{\mymaximum}[1]{\mathop{\textrm{maximum}}_{#1}}    
\newcommand{\mymin}[1]{\mathop{\textrm{minimize}}_{#1}}
\newcommand{\mymax}[1]{\mathop{\textrm{maximize}}_{#1}}
\newcommand{\mymins}[1]{\mathop{\textrm{min.}}_{#1}}
\newcommand{\mymaxs}[1]{\mathop{\textrm{max.}}_{#1}}  
\newcommand{\myargmin}[1]{\mathop{\textrm{argmin}}_{#1}} 
\newcommand{\myargmax}[1]{\mathop{\textrm{argmax}}_{#1}} 
\newcommand{\myst}{\textrm{s.t. }}

\newcommand{\denselist}{\itemsep -1pt}
\newcommand{\sparselist}{\itemsep 1pt}

\definecolor{pink}{rgb}{0.9,0.5,0.5}
\definecolor{purple}{rgb}{0.5, 0.4, 0.8}   
\definecolor{gray}{rgb}{0.3, 0.3, 0.3}
\definecolor{mygreen}{rgb}{0.2, 0.6, 0.2}

\newcommand{\cyan}[1]{\textcolor{cyan}{#1}}
\newcommand{\red}[1]{\textcolor{red}{#1}}  
\newcommand{\blue}[1]{\textcolor{blue}{#1}}
\newcommand{\magenta}[1]{\textcolor{magenta}{#1}}
\newcommand{\pink}[1]{\textcolor{pink}{#1}}
\newcommand{\green}[1]{\textcolor{green}{#1}} 
\newcommand{\gray}[1]{\textcolor{gray}{#1}}    
\newcommand{\mygreen}[1]{\textcolor{mygreen}{#1}}    
\newcommand{\purple}[1]{\textcolor{purple}{#1}}       

\definecolor{greena}{rgb}{0.4, 0.5, 0.1}
\newcommand{\greena}[1]{\textcolor{greena}{#1}}

\definecolor{bluea}{rgb}{0, 0.4, 0.6}
\newcommand{\bluea}[1]{\textcolor{bluea}{#1}}
\definecolor{reda}{rgb}{0.6, 0.2, 0.1}
\newcommand{\reda}[1]{\textcolor{reda}{#1}}

\def\changemargin#1#2{\list{}{\rightmargin#2\leftmargin#1}\item[]}
\let\endchangemargin=\endlist
                                               
\newcommand{\cm}[1]{}

\newcommand{\mhoai}[1]{{\color{magenta}\textbf{[Hoai: #1]}}}
\newcommand{\phuc}[1]{{\color{blue}\textbf{[Phuc: #1]}}}

\newcommand{\mtodo}[1]{{\color{red}$\blacksquare$\textbf{[TODO: #1]}}}
\newcommand{\myheading}[1]{\vspace{1ex}\noindent \textbf{#1}}
\newcommand{\htimesw}[2]{\mbox{$#1$$\times$$#2$}}

\newcommand{\etal}[0]{\textsl{et al.}}

% The following are useful for creating homework or exams

\newif\ifshowsolution
%\showsolutionfalse
\showsolutiontrue

\ifshowsolution  
\newcommand{\Comment}[1]{\paragraph{\bf $\bigstar $ COMMENT:} {\sf #1} \bigskip}
\newcommand{\Solution}[2]{\paragraph{\bf $\bigstar $ SOLUTION:} {\sf #2} }
\newcommand{\Mistake}[2]{\paragraph{\bf $\blacksquare$ COMMON MISTAKE #1:} {\sf #2} \bigskip}
\else
\newcommand{\Solution}[2]{\vspace{#1}}
\fi

\newcommand{\truefalse}{
\begin{enumerate}
	\item True
	\item False
\end{enumerate}
}

\newcommand{\yesno}{
\begin{enumerate}
	\item Yes
	\item No
\end{enumerate}
}

\newcommand{\Sref}[1]{Sec.~\ref{#1}}
\newcommand{\Eref}[1]{Eq.~(\ref{#1})}
\newcommand{\Fref}[1]{Fig.~\ref{#1}}
\newcommand{\Tref}[1]{Table~\ref{#1}}

\section{Introduction}

Maps, representing one of the earliest forms of images, are deeply significant to our understanding of the environment and our place within it. They encapsulate more than just geographical information; maps are instilled with cultural, political, and historical meanings, making them a rich subject for analysis. Building a computer vision algorithm for map analysis is crucial as it enables the automatic extraction and interpretation of these layers of data embedded within maps. Such technology not only enhances our ability to understand historical changes and cultural insights but also aids in real-time detection and prevention of malicious activities. By leveraging computer vision for map analysis, we can unlock a more nuanced and comprehensive understanding of the multifaceted information that maps provide, ensuring that this valuable resource can be utilized to its fullest potential in various fields.

\begin{figure}[h]
\centerline{\includegraphics[width=0.8\linewidth]{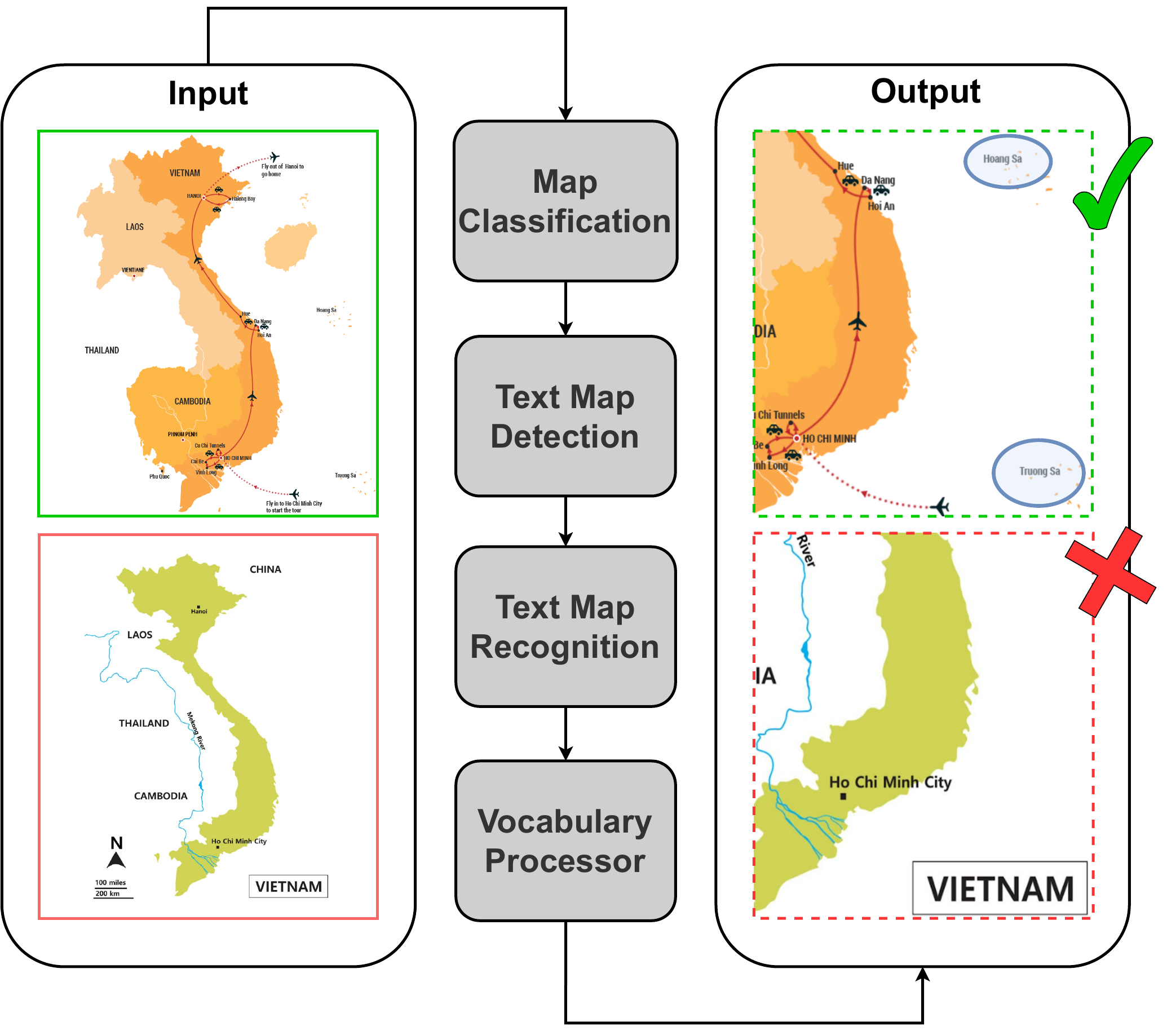}}
\caption{Our proposed model, pre-trained on high-quality VinMap datasets, demonstrates the capability to recognize the Vietnam map and determine whether it includes the Hoang Sa or Truong Sa regions from multi-resolution input map images.}
\label{fig:teaser}
\end{figure}

One particular computer vision technology that is potentially beneficial is the one capable of identifying maps that depict specific regions and feature landmarks with designated names. This technology is crucial in a variety of scenarios, especially when it involves navigating through vast archives of historical documents or continually scanning produced content. These capabilities are indispensable for historical research, such as determining when a city or region was first referred to by a specific name---a key to understanding geopolitical shifts, including the adoption of new political boundaries or name changes due to evolving political dynamics. For example, this technology can pinpoint the time when the ancient city of Alexandria was first labeled as such on maps, as well as when its inhabitants began to recognize and embrace this name. Moreover, this technology could be pivotal in demystifying mythical places such as Atlantis, analyzing cartographic records across different eras to trace the development of its legend and its impact on cultural narratives.

Having mentioned the potential use of computer vision for automatic map analysis, its efficacy for identifying maps that depict a specific region with landmarks bearing designated names remains uncertain. This complexity arises from the need to tackle three non-trivial subtasks: 1) distinguishing maps from non-map images; 2) verifying that the map in question actually represents the region of interest; and 3) confirming the presence or absence of a specific landmark name on the map. These subtasks are challenging due to the diverse styles and methods used in map creation. In the first subtask, although differentiating maps from other types of images might seem straightforward given the current advancement in computer vision and transfer learning, achieving perfect accuracy is complicated by the fact that hand-drawn maps can closely resemble other artistic illustrations. The second subtask, determining whether or not a map contains the targeted area, is also challenging because maps of the same locale can vary significantly in appearance based on their intended content and creation process. Conversely, maps of different regions may share stylistic elements, making them misleadingly alike. The final subtask involves text spotting and recognition, which is rendered difficult by the variety of text presentations, ranging from neatly printed to cursive handwriting, and including text set in vertical, slanted, or curved orientations.

To assess the efficacy of computer vision for map analysis, this paper develops a method and evaluates its performance through a particular scenario. We consider the challenge of sifting through an extensive collection of images to identify maps that either depict Vietnam in its entirety or include segments of it. Furthermore, we consider the nuanced task of determining if such a map specifically excludes the contested islands Hoang Sa (Paracel) and Truong Sa (Spratly), which are at the center of international territorial disputes. While acknowledging the inherent political sensitivity of this issue, our study deliberately concentrates on the technical aspects. The selected problem serves as an exemplar for the actual demands of automated map analysis, encompassing all the subtasks previously outlined and providing a comprehensive test case for our developed method.

%To figure out the answer to the aforemenioned question of the efficacy of computer vision for map analysis, in this paper, we develop such a method and evaluate its performance on a specific use case. Specifically, we consider the task of scanning a huge colleciton of images to detect maps of Vietnam or contain parts of Vietnam. We are also consider the task of deteciting the maps of Vietnam that do not show the two sets of islands called Hoang Sa and Troung Sa. These islands are subject of international claims and disputes, so the problem does have some potentially political significant itself. However, we set this aside, and we focus on the technical problem here. This problem is good good representative use case of the actually need for automatic map analysis, and it is also fully contain the subtasks that we need to address as mentioned above. 

%Our approach is built upon cutting-edge computer vision methodologies. Initially, we classify whether the input image is a Vietnam map. Subsequently, we execute Text Detection and Recognition on the map to identify and interpret each text segment, aiming to grasp the map's semantic content. Following this, Map Vocab Matching establishes a similarity matrix between the predicted texts and a predefined policy to verify the presence of the two islands on the map

Our approach leverages state-of-the-art computer vision techniques presented in Fig. \ref{fig:teaser}. Initially, using a Convolutional Neural Network \cite{tan2020efficientnet, liu2022convnet} and transfer learning \cite{zhuang2020transfer}, it determines whether an input image is a map of Vietnam. If the image passes this initial test, we proceed to detect and recognize all text present on the map. Subsequently, each instance of recognized text is cross-referenced with acceptable variations of the names Truong Sa and Hoang Sa to confirm whether these islands are depicted on the map.

To train and evaluate the performance of our method, we have assembled a dataset comprising a variety of map images of Vietnam, either in entirety or in sections, which will be referred to as the \textbf{Vi}et\textbf{n}am Map or VinMap collection. This dataset includes a range of maps obtained from varied geographic sources, featuring inscriptions in either Vietnamese or English. The collection is organized with labels that confirm the map's depiction of significant territories, the Hoang Sa and Truong Sa islands. Additionally, it is enhanced with box annotations that mark the precise locations of text related to the Hoang Sa and Truong Sa, aiding in the fine-tuning of text recognition processes.

Experiments performed on the VinMap dataset reveal that our method can detect maps of Vietnam that exclude the Truong Sa and Hoang Sa islands with precision and recall values of 78.51\% and 93.87\%, respectively. These findings illustrate both the strengths and limitations of the approach. On the positive side, the algorithm performs well enough to be considered for practical applications, particularly in scenarios that involve large-scale map scanning where some level of human verification is acceptable. However, the results are not flawless, indicating a clear need for continued research and improvement in this field.

%Experiments on the VinMap dataset show that our developed method achieves an F1-score of $85.30$ for the task of detecting Vietnam map that does not contain either Truong Sa or Hoang Sa. The precision and recall are above 80\% and 90\% respectively. These results underscore two sides of this problem. On the positive side, the algorithm is good enough for their potential use in some applications, especially for applications that involve large scale scanning of maps and some human verification is acceptable. On the negative side, the result is far from perfect, highlighting the need for further research and development in this area. 

%These results underscore the significant challenges presented within the VinMap dataset, where Vietnamese maps are often characterized by blurriness, low resolution, and diverse styles. These factors make semantic information extraction from recent map analysis methods challenging, particularly in recognizing geographic regions of interest. Furthermore, our meticulously developed pipeline showcases its dependability on the demanding dataset, laying the groundwork for future research endeavors.

In short, the contributions of this paper are threefold. First, we introduce and explore a new avenue in map analysis, an area ripe for investigation with significant potential impact. Second, we have developed a complete program harnessing cutting-edge computer vision technologies, and we examine its effectiveness through a specific use case. Our experimental results reveal satisfactory performance by our method, underscoring the value of computer vision while also highlighting avenues for further enhancements. Lastly, we introduce the VinMap dataset, a comprehensive collection of thousands of annotated map images, which serves not only as the foundation for developing and testing our approach but may also be instrumental for future endeavors in map analysis tasks.

\section{Related Works}
% Prior work? mentioning text-guided dataset, ICDAR, etc
% Prior Dataset
\subsection{Map Classification}
% zhu2017similarity
Previous research \cite{li2002quantitative,fritz2005comparison} addressed the map-matching problem and proposed solutions at a local image scale, primarily utilizing methods that aggregate observed elementary data similarities rather than deep learning frameworks. In contrast, deepMap dataset was \cite{zhou2018deep} introduced to explore map classification using deep learning techniques. They employed a straightforward deep convolutional neural network architecture, which led to significant improvements compared to heuristic approaches. Subsequent studies \cite{arundel2020geonat,li2023computational} have further advanced this field by leveraging deep learning models to extract deep-level features for multi-resolution maps. However, recent practice favors pre-training the model on datasets like \cite{russakovsky2015imagenet} before fine-tuning it on a specific dataset, as it yields more promising results, becomes a standard approach.  
\subsection{Text Detection}
Text detection is an important research problem and it has received much research attention. 
Earlier studies \cite{adak2013unsupervised, grover2009text} used machine learning clustering-based algorithms to extract text from background images. While these methods are relatively straightforward, they tend to achieve inferior performance. Recently, deep learning-based approaches \cite{liao2022real, liao2020real} have shown much better performance in text detection tasks. In our work, we propose to fine-tune a detection model that has been pre-trained on a public dataset such as \cite{wu2023icdar, nguyendictguide} to our dataset to direct its attention and adapt it to the task of detecting text on maps, especially text for the landmark names of the two sets of islands. 

%Fine-tuning the text detection model on a specific dataset directs its attention towards key regions, particularly in this context, the two islands.

\subsection{Text Recognition}
% cirecsan2010deep
% driess2023palm
Previous work on Text recognition or Optical Character Recognition (OCR) \cite{al2009handwriting} typically employed simple Neural Networks to perform logistic regression for preset characters. However, due to their simplicity, these models can only recognize a single character at a time. In contrast, recent Transformer-based methods \cite{li2023trocr, lyu2022maskocr} are fast, scalable, and patch-based, achieving promising results by processing a text region as a whole query-able feature vector. Particularly noteworthy is the fine-tuning of OCR models on specific language datasets such as \cite{nguyendictguide} from pre-trained ENG text recognition datasets like \cite{karatzas2015icdar, cheng2021icdar}, enabling them to become multilingual.
\section{Task and Dataset}
This section introduces a new challenge in map understanding and offers detailed statistics for the newly proposed VinMap dataset.
\subsection{Task definition}
\label{sec:problem}
We focus on the task of scanning images to identify Vietnam maps that do not include Hoang Sa (Paracel) and Truong Sa (Spratly). We consider both English and Vietnamese text descriptions of the two islands. We frame this as a detection problem, where the positive class comprises Vietnam maps that exclude both Hoang Sa (Paracel) and Truong Sa (Spratly). All other cases are considered negative, including non-map images, non-Vietnam maps, or Vietnam maps that contain either Hoang Sa (Paracel) or Truong Sa (Spratly). 

%\phuc{The current description is correct (either)}\mhoai{Double check either or neither. A Vietnam map is considered negative if it contains either or neither the island? There is some inconsistency with the description of the datafset below}

% Given an input map image $I$, we first determine if it is a Vietnam map or not, and filter out images that are not Vietnam maps. For a Vietnam map, we then run text 

% \begin{itemize}
%     \item \textbf{Not Vietnam map}. Discard as we only focus on Vietnam map.
%     \item \textbf{Vietnam map}. We then ascertain the validity of the map by verifying the inclusion of both the Hoang Sa (Paracel) and Truong Sa (Spratly) islands.
% \end{itemize}

% We then regard the final prediction of the map:
% \begin{itemize}
%     \item \textbf{Positive}: Vietnam map and do not contain both of the islands.
%     \item \textbf{Negative}: (not Vietnam map) or (VN map and contain either the islands).
% \end{itemize}
% We then address the map's validity by framing it as a Text Detection and an Optical Character Recognition (OCR) problem.

\subsection{VinMap Dataset}

\begin{figure*}[h]
\centerline{\includegraphics[width=13cm]{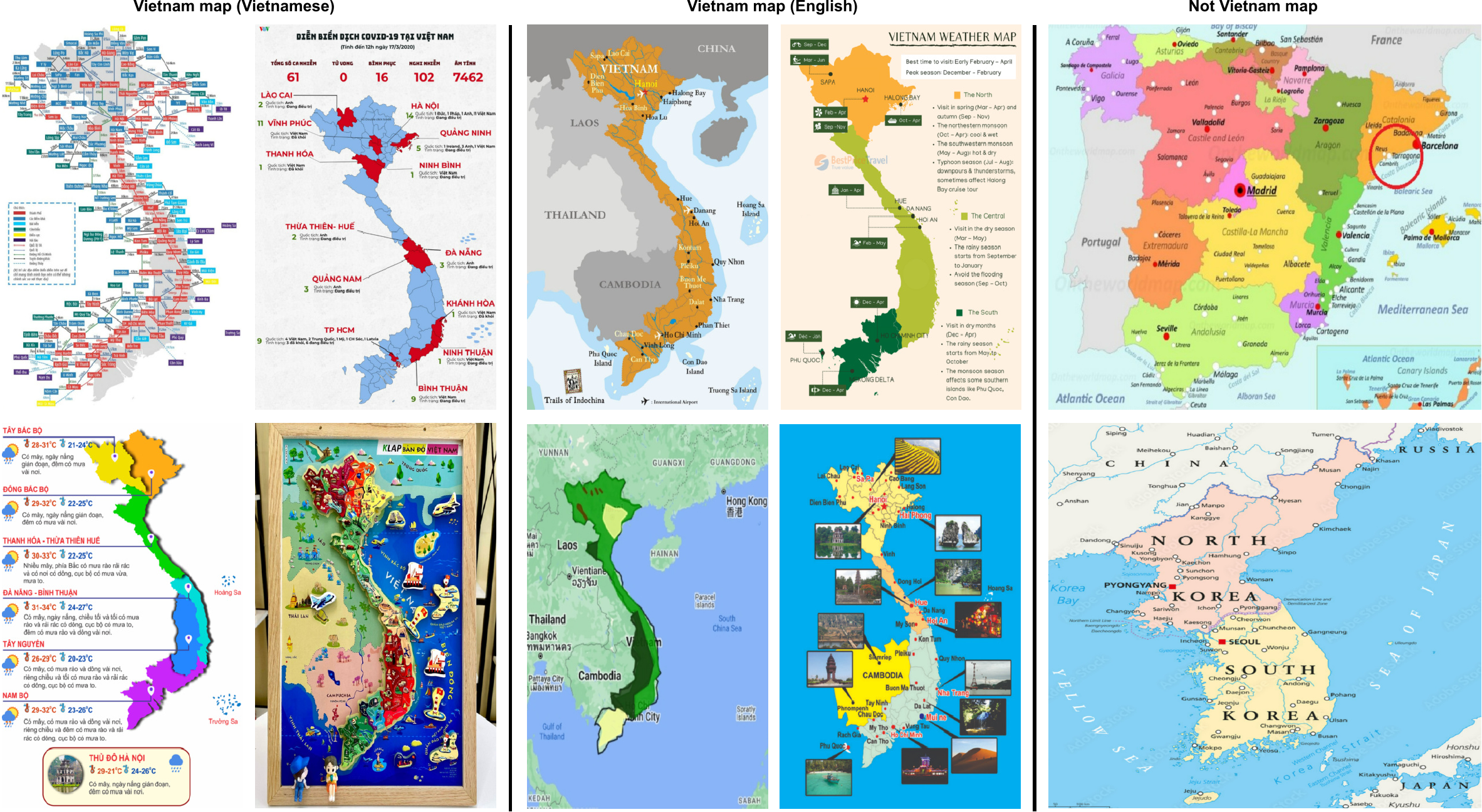}}
\caption{The VinMap dataset comprises high-quality images in both English and Vietnamese. Tailored specifically for Vietnam, the Vietnam map set encompasses maps depicting various contexts of Vietnam, whereas the Not Vietnam map set comprises map images from diverse countries and regions.}
\label{fig:dataset_quali}
\end{figure*}

\begin{figure}[h!]
\centerline{\includegraphics[width=0.75\linewidth]{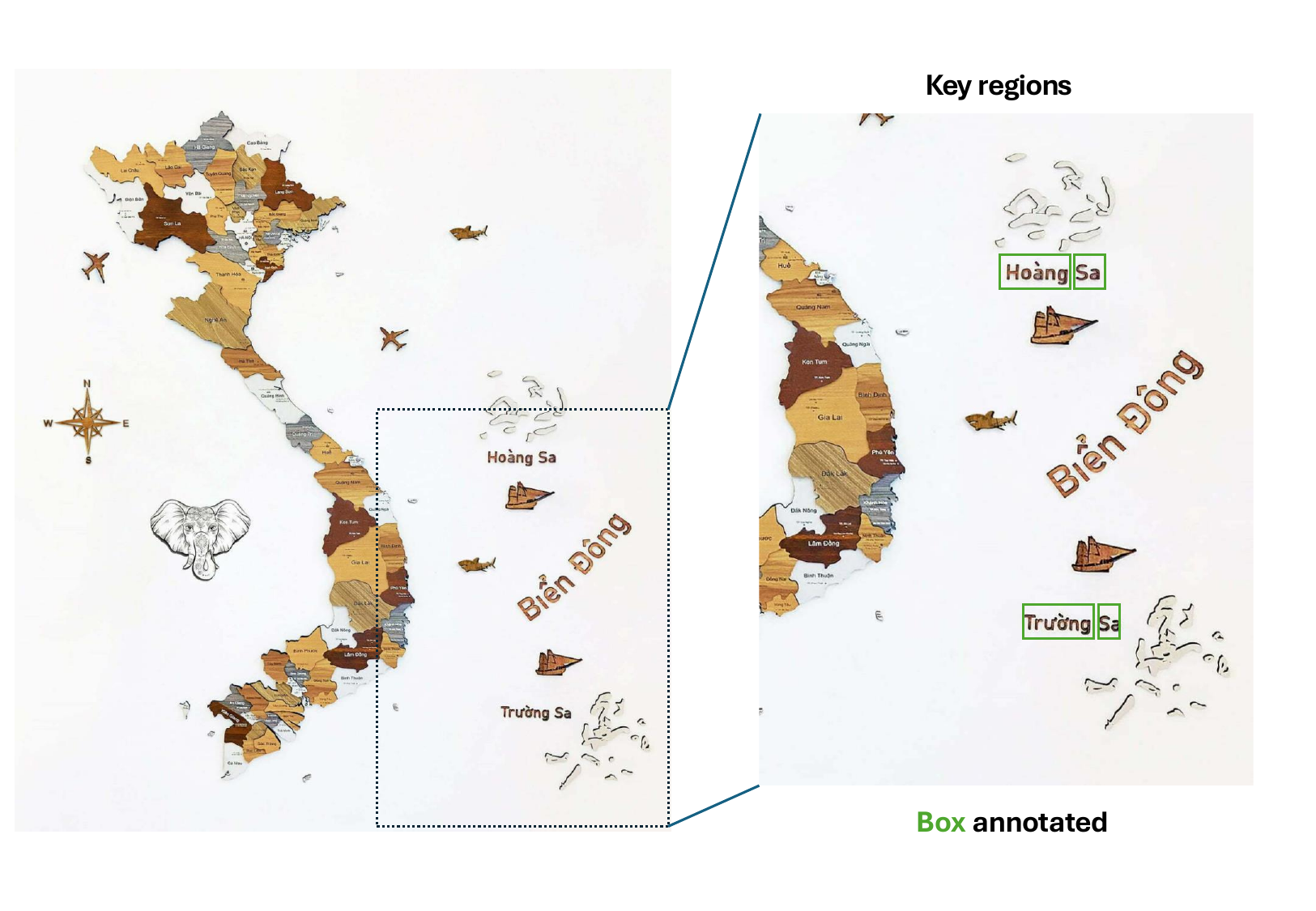}}
\caption{Annotation visualization of the Vietnam map image and our provided box annotation for regions of interest.\label{fig:anno_quali}}
\end{figure}

The VinMap dataset comprises a total of 6,858 images with diverse resolutions. Among these, 2,000 images are non-map images, 2,777 maps do not depict Vietnam, and 1,002 maps represent Vietnam and include either the Truong Sa or Hoang Sa islands (866 maps are in Vietnamese, and 136 maps are not in Vietnamese). There are 1,079 maps of Vietnam that do not contain both the Truong Sa and Hoang Sa islands (291 maps are in Vietnamese, and 788 maps are not in Vietnamese). Vietnam maps encompass various geographic regions, yet to instruct vision models to prioritize specific map areas such as Truong Sa and Hoang Sa, adhering to governmental regulations, VinMap offers box annotations for every Vietnam map containing both the Truong Sa and Hoang Sa islands. This meticulous annotation process establishes the groundwork for advancing map analysis research in Vietnam. Box annotations are depicted in Fig.~\ref{fig:anno_quali}. Table~\ref{tab:vinmap_stats} summarizes the statistics of the VinMap dataset. Some images of VinMap are shown in Fig.~\ref{fig:dataset_quali}. The dataset presents several advantages, introducing more challenging aspects than previous map datasets.

\setlength{\tabcolsep}{3pt}
\begin{table}[t]
    \centering
    \begin{tabular}{llrrrc}
    \toprule 
       Type of images  & Language  & \#Train & \#Test &  Total & Annotation? \\
       \midrule 
         Not maps & Mixed  & {1400} &  {600} & 2,000 & \xmark \\
         \midrule 
         Not Vietnam maps & Mixed & {1944} & {833 }& 2,777 & \xmark  \\
         \midrule 
         \multirow{ 3}{15ex}{Vietnam maps containing \hspace{2ex} (TS or HS)} & Vietnamese &  {606} & {260} & 866 & \multirow{3}{*}{\cmark}  \\
          & English & {95} & {41} & 136 \\
          \cmidrule{2-5}
          & Sub-total & {701} & {301} & 1,002 \\
          \midrule 
         \multirow{3}{15ex}{Vietnam maps not containing (TS and HS)} & Vietnamese & {204} & {87} & 291 & \multirow{3}{*}{\xmark}\\
          & English & {552} & {236} & 788 \\
          \cmidrule{2-5}
          & Sub-total & {756} & {323} & 1,079 \\          
          \midrule
          Total & & 4,801& 2,057 & 6,858 \\
    \bottomrule 
    \end{tabular}
    \caption{Statistics of the VinMap dataset. There is a total of 6,858 images, divided into disjoint training and testing subsets of 4801 and 2057 images, respectively.}
    \label{tab:vinmap_stats}
\end{table}

% 1.ContainingTSHS/Vietnamese 866
% 1.ContainingTSHS/NotVietnamese 136
% 2.NotcontainingTSHS/Vietnamese 291
% 2.NotcontainingTSHS/NotVietnamese 788
% 3.MapnotcontainingVN 2777
\section{Experiments}
\subsection{Proposed Method}
\begin{figure*}[h]
\centerline{\includegraphics[width=12cm]{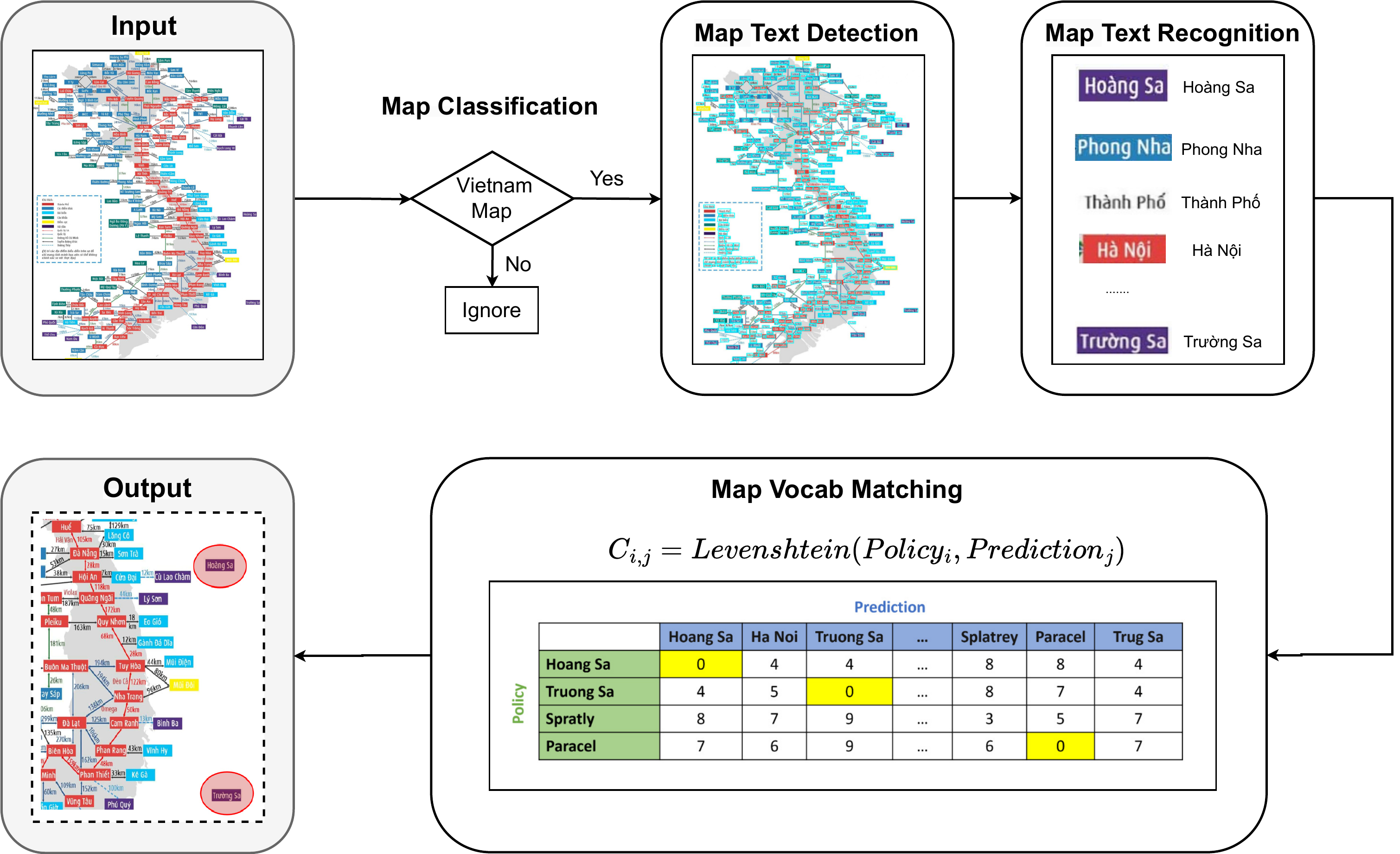}}
\caption{The proposed pipeline for VinMap comprises four stages. Initially, the Map Classification module determines whether a given input image is a Vietnam map. If affirmative, the Map Text Detection module identifies all text regions within the map using computer vision techniques. Subsequently, the Map Text Recognition module scans these detected text regions and predicts the corresponding texts using an OCR model. Finally, the predicted texts are compared with a predefined policy using Levenshtein distance by the Map Vocab Matching to ascertain whether the input map image contains key regions.}
\label{fig:baseline}
\end{figure*}

This section describes the proposed method, which consists of several steps: map classification, text detection, text recognition, and vocab matching; depicted in Fig. \ref{fig:baseline}

\myheading{Map classification.} Our objective is to categorize map images into two groups: those that depict Vietnam and those that do not. To accomplish this, we utilize the EfficientNet-B4 classification model \cite{tan2020efficientnet}, modifying the final classification layer to output only two categories instead of the original 1,000. The remaining layers are initialized with weights pre-trained on the ImageNet dataset \cite{russakovsky2015imagenet}. Training is conducted on the training set of VinMap comprising 4,801 images, with 3,344 non-Vietnam maps (1,400 non-map images, 1,944 maps depicting regions other than Vietnam) and 1,457 maps of Vietnam. We employ Cross-Entropy Loss over 100 epochs, with a batch size of 4 and a learning rate of 0.1. Additionally, we apply random crop-flip augmentations to enhance training data diversity.
% \mhoai{Do you train on all images? Don't you have separate test data?} \phuc{as mentioned, I use all imgs}

\myheading{Text Detection.} 
Our objective is to spot semantic text regions within map images depicting Vietnam, with a particular focus on the key regions of the two islands. Initially, we utilize DBNet \cite{liao2020real}, pre-trained on the English ICDAR2015 dataset \cite{karatzas2015icdar}. We adopt a two-step training approach.
In the first step, we aim to adapt the model to recognize Vietnamese text regions. To achieve this, we fine-tune the model on 33,000 Vietnamese text instances from 1,200 training images from the VinText dataset \cite{nguyendictguide}.
In the second step, we refine the model to specifically focus on the key regions (text regions of the two islands). For this stage, we fine-tune the model using the provided training set VinMap box annotations of the two islands within 701 maps.
Throughout both stages, we utilize the ResNet50 backbone and employ combination probability, binary, and threshold map losses \cite{liao2020real}, with a batch size of 2 and a learning rate of 0.001. Additionally, we incorporate random crop and rotate augmentations to better accommodate map images.
% \mhoai{Do you separate train/test split? It is not clear to me.}
% \phuc{as mentioned, I did not split train/test. Is it OK to declare that we don't split the data, anh Hoai?}

\myheading{Text Recognition.}
The objective is to comprehend the semantic information extracted from the detected text regions. To accomplish this, we utilize the open-source VietOCR \footnote{https://github.com/pbcquoc/vietocr}, which is built upon the Transformer OCR architecture \cite{li2023trocr}. This VietOCR tool has been pre-trained on an extensive dataset comprising over 10 million synthetic, handwritten, and scanned images. Since the pre-trained OCR model performed effectively on both Vietnamese and English text, we did not finetune it further. 

\myheading{Vocab Matching.} 
The objective is to align predicted text instances with a predefined vocabulary policy. This involves calculating the Levenshtein distance \footnote{https://en.wikipedia.org/wiki/Levenshtein\_distance} between the known vocabulary and the text predicted by the OCR model described in Fig. \ref{fig:baseline}. If the distance is smaller than a threshold value, denoted as $\lambda = 2$, the predicted text region is considered a match with the key text regions—in our case, the two islands.

\subsection{Evaluation protocol}
We regard the map's final prediction according to Section \ref{sec:problem}. To evaluate the proposed pipeline on the VinMap dataset, we consider:
\begin{itemize}
    \item \textbf{Precision}: measures the accuracy of the positive predictions. It answers the question: ``Of all the images predicted as Vietnam maps that do not contain Truong Sa and Hoang Sa, how many were actually maps that do not contain Truong Sa and Hoang Sa?''
    \item \textbf{Recall}: measures the ability of the model to find all the Vietnam map that do not contain Truong Sa and Hoang Sa. It answers the question: "Of all the actual Vietnam maps that do not contain Truong Sa and Hoang Sa, how many were correctly detected?"
    \item \textbf{F1-Score}: The harmonic mean of precision and recall provides a balanced measure that considers both precision and recall equally: $F_1 = \frac{2 \times Precision \times Recall}{ Precision + Recall}$
\end{itemize}

% \mhoai{Came here. Will resume editing later}
\subsection{Results Analysis}

% (
% - Raw Image Classification baseline pretrained on Map set containing HS and TS; the others
% - Main table
% )

% (- Ablation Study)

\begin{table}[htp]
\centering
\caption{The comparison of the proposed pipeline and the naive image classification approach}
\label{tab:result}
\renewcommand{\arraystretch}{1.5}
\begin{tabular}{l|c|c|c|c}
\toprule
\textbf{Method} & \multicolumn{1}{c|}{\textbf{Test setting}}&
  \multicolumn{1}{c|}{\textbf{Precision}} &
  \multicolumn{1}{c|}{\textbf{Recall}} &
  \multicolumn{1}{c}{\textbf{F1-Score}} \\ \hline
Raw Image Classification & ENG-VN & 39.25 & 53.84 & 45.40 \\ \hline
VinMap & ENG & 93.12 & 95.91 & 94.49 \\
VinMap  & VN & 65.53 & 91.28 & 76.29 \\
VinMap & ENG-VN & 78.51 & 93.87 & 85.51 \\ \bottomrule
\end{tabular}
\end{table}

\begin{table}[htp]
\centering
\caption{Results on the related tasks}
\label{tab:tasks}
\renewcommand{\arraystretch}{1.5}
\begin{tabular}{l|c}
\toprule
\textbf{Task} &
\textbf{AP} \\ \midrule
detect map from ALL images  & 99.6 \\
detect VN maps from ALL images & 97.52  \\
detect VN maps not containing (HS and TS) from ALL images & 75.21 \\ \bottomrule
\end{tabular}
\end{table}

\begin{table}[htp]
\centering
\caption{Ablation Study of the proposed pipeline}
\label{tab:ablate}
\renewcommand{\arraystretch}{1.5}
\begin{tabular}{l|c|c|c|c}
  \multicolumn{1}{c}{} &
  \multicolumn{1}{c}{} &
  \multicolumn{1}{c}{} &
  \multicolumn{1}{c}{}\\ \toprule
\multirow{2}{*}{\textbf{Classification}} & {Pretrained ImageNet} & x & x &  \\  
& Non-Map images included & x &  & x\\ \hline
F1-Score & & 85.51 & 62.34 &  55.62 \\ \bottomrule
\toprule

\multirow{3}{*}{\textbf{Text Detection}} & Pretrained ICDAR2015 & x & x & x\\
& Fine-tune VinText  & x & x &   \\
& Fine-tune VinMap box & x &  &   \\ \hline
F1-Score & & 85.51 & 72.34 & 63.15 \\ \bottomrule
\toprule
{\textbf{Vocab Matching}} & {$\lambda$}  & 5  &  2 & 1 \\ \hline
F1-Score & & 63.44 & 85.51 & 77.23 \\ \bottomrule
\end{tabular}
\end{table}

We present detailed quantitative results for VinMap using both the proposed pipeline and the Raw Image Classification pipeline, as illustrated in Table \ref{tab:result}. In the Raw Image Classification setting, we train a binary classification EfficientNet-B4 model on the VinMap training set following the policy outlined in Section \ref{sec:problem}. The direct approach yields only a 45.40 \% F1-Score on the English-Vietnamese test set, indicating significant challenges posed by the VinMap dataset. Conversely, our proposed pipeline demonstrates substantial improvement across three evaluation metrics for the dataset. Specifically, in the case of the Vietnamese maps test set, our method experiences a notable drop from the English maps test set by 18.2\% F1-Score and 27.59\% Precision score, while maintaining a Recall rate of over 90\%. The decrease in performance can be ascribed to the challenges associated with identifying five distinct diacritics in Vietnamese text images within the Map Text Recognition module. The detected text regions (see Fig. \ref{fig:baseline}), cropped from map images, frequently display blurriness and noise owing to the low resolution of the maps.

We delve deeper into related tasks, presenting the recorded Average Precision (AP) scores outlined in Table \ref{tab:tasks}. The initial experiment entails utilizing our Map Classification model. In this experiment, the classification model is trained on a dataset comprising 1,400 non-map images and the remaining 3,401 map images, achieving an AP score of 99.6\%. The last two experiments are based on the classification model and the final predictions derived from our method.

We explore various configurations of the proposed method, as outlined in Table \ref{tab:ablate}. Each module is studied independently, with one module being investigated while the others remain at their default settings. Evaluation F1-Score is conducted on the overall pipeline's final prediction.

In the Map Classification module, we conduct ablations on the EfficientNet-B4 model by examining whether to utilize the pre-trained backbone on ImageNet1000 \cite{russakovsky2015imagenet} and whether to include the Non-Map images set during training. Results indicate that utilizing the EfficientNet-B4 model pre-trained from ImageNet1000 significantly improves the understanding of map images, as evidenced by a performance drop to 55.62\% F1-Score when not using it. Additionally, including 1,400 Non-Map images during the training process of the model pre-trained from ImageNet1000, boosts overall performance by 85.51 F1-Score, enabling the model's capability to differentiate between real-world and map images.

In the ablation of the Map Text Detection module, we study refining the model's performance across various datasets. Primarily, ensuring the model can effectively detect English text regions necessitates pretraining it on ICDAR2015 \cite{karatzas2015icdar}. Moreover, enhancing the model's detection capabilities across both English and Vietnamese text entails fine-tuning the detection module on VinText, resulting in an improved overall F1-Score to 72.34 from only 63.15 when using only the pre-trained ICDAR20215. Additionally, to guide the detection model to focus on specific regions on the map, such as the two islands mentioned in this context, we further fine-tuned the model using the proposed VinMap box annotation set, leading to a notable surge in F1-Score to 85.51. This validates that our proposed annotation set on VinMap effectively directs the detection toward identifying regions of interest on map images.

We examine the impact of $\lambda$ on the proposed Vocab Matching Algorithm. Decreasing $\lambda$ entails a stricter adherence to matching the predicted text regions from our Map Text Recognition module with the specified policy terms (in our case, "Hoang Sa", "Truong Sa", "Spratly", "Paracel"). As illustrated in Table \ref{tab:ablate}, the highest F1-Score of 85.51 is achieved when $\lambda=2$. However, adjusting $\lambda=1$ inadvertently disregards near-perfect text predictions such as "Trung Sa", "Hoag Sa", "Spatly", "Parcl", etc., leading to misinterpretations of the map images. Conversely, relaxing $\lambda=5$ significantly impacts performance with predicted regions like "Trung Son", "Ha Noi", etc.

\section{Conclusion}
Conclusively, we present a pioneering endeavor in geographic map comprehension through the introduction of the challenging dataset VinMap, comprising meticulously annotated map images. Furthermore, we establish a resilient method utilizing contemporary computer vision methodologies to scrutinize the dataset, thus laying the groundwork for forthcoming explorations in map analysis.

\myheading{Acknowledgement.} We sincerely thank the MIC-VN team for data crawling and labeling. We thank Mr. Que Nguyen and his team for their support in testing and deployment.

\printbibliography
\end{document}